\pdfoutput=1

\documentclass[11pt]{article}

\usepackage[]{acl}

\usepackage{times}
\usepackage{latexsym}
\usepackage{arydshln}

\usepackage[T1]{fontenc}

\usepackage[utf8]{inputenc}

\usepackage{microtype}
\usepackage{comment}
\usepackage{amsmath}
\usepackage{inconsolata}
\usepackage{multirow}
\usepackage{makecell}
\usepackage{graphicx}
\usepackage{booktabs}
\usepackage{subfig}
\usepackage{enumitem}
\title{Knowledge-Aware Query Expansion with Large Language Models for Textual and Relational Retrieval}

\author{Yu Xia$^{1}$ \qquad Junda Wu$^{1}$ \qquad Sungchul Kim$^{2}$ \qquad Tong Yu$^{2}$ \\ {\bf Ryan A. Rossi$^{2}$ \quad\; Haoliang Wang$^{2}$ \quad\; Julian McAuley$^{1}$} \\ 
$^{1}$University of California San Diego \qquad 
$^{2}$Adobe Research \qquad\\
\texttt{\{yux078, juw069, jmcauley\}@ucsd.edu} \\ \texttt{\{sukim, tyu, ryrossi, hawang\}@adobe.com}}

\begin{document}
\maketitle
\begin{abstract}
Large language models (LLMs) have been used to generate query expansions augmenting original queries for improving information search. 
Recent studies also explore providing LLMs with initial retrieval results to generate query expansions more grounded to document corpus.
However, these methods mostly focus on enhancing textual similarities between search queries and target documents, overlooking document relations.  
For queries like ``Find me a highly rated camera for wildlife photography compatible with my Nikon F-Mount lenses'',
existing methods may generate expansions that are semantically similar but structurally unrelated to user intents.
To handle such semi-structured queries with both textual and relational requirements, in this paper we propose a knowledge-aware query expansion framework, augmenting LLMs with structured document relations from knowledge graph (KG).
To further address the limitation of entity-based scoring in existing KG-based methods,
we leverage document texts as rich KG node representations and use document-based relation filtering for our \textbf{K}nowledge-\textbf{A}ware \textbf{R}etrieval (KAR).
Extensive experiments on three datasets of diverse domains show the advantages of our method compared against state-of-the-art baselines on textual and relational semi-structured retrieval.
\end{abstract}

\section{Introduction}
Large language models (LLMs) have been utilized to expand original queries with additional contexts, capturing similar semantics of target documents and hence improving retrieval performance \cite{gao-etal-2023-precise, wang-etal-2023-query2doc}.
While direct generations of LLMs introduce problems such as hallucination, out-dated information, and lack of domain knowledge, recent methods \cite{jagerman2023query, lei-etal-2024-corpus, shen-etal-2024-retrieval} explore augmenting LLMs with initial retrievals as contexts, e.g., pseudo relevance feedback (PRF) \cite{lv2010positional, li2022improving}, to generate expansions more grounded to domain-specific document corpus.

\begin{figure}[!t]
    \centering
    \includegraphics[width=0.98\linewidth]{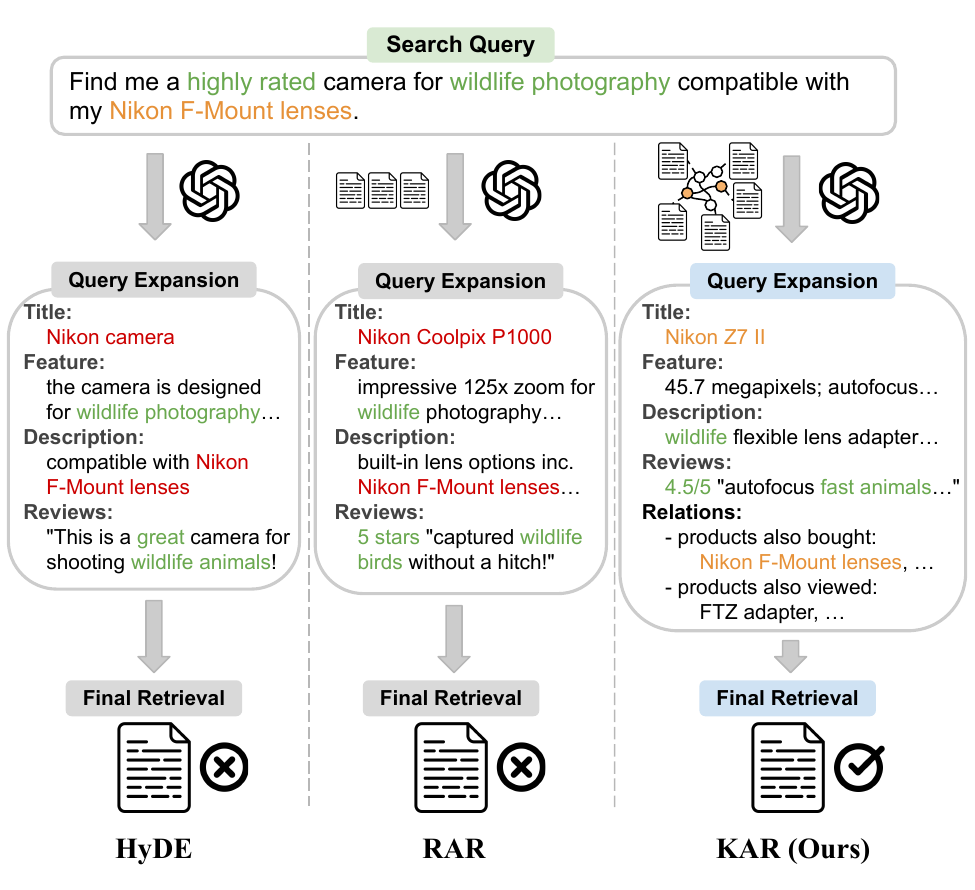}
    \caption{Example query expansions generated by HyDE \cite{gao-etal-2023-precise}, RAR \cite{shen-etal-2024-retrieval}, and KAR (Ours) given a semi-structured product search query with both \textcolor[rgb]{0.39, 0.61, 0.29}{textual} and 
  \textcolor[rgb]{0.89, 0.56, 0.22}{relational} requirements \cite{wu2024stark}. While HyDE and RAR enrich the textual information, e.g., ``wildlife'' and ``highly rated'', they make up \textcolor[rgb]{0.8, 0, 0}{incorrect} document relations, e.g., compatibility of ``Nikon Coolpix P1000'' with ``F-Mount lenses''. In contrast, our KAR utilizes document relations from knowledge graph, e.g., customers bought ``Nikon Z7 II'' and ``F-Mount lenses'' together, to generate semantically similarly and structurally related query expansions. 
    }
    \label{fig:intro}
    \vspace{-1em}
\end{figure}

Though effective, existing methods mostly focus on enhancing the semantic similarities between expanded queries and target document texts.
However, in real-world search scenarios, besides textual descriptions, documents are often inter-connected with certain types of relations \cite{talmor-berant-2018-web, cao-etal-2022-kqa}.
Both textual and relational details are often queried by users in semi-structured manner \cite{wu2024stark, patel2024lotus, wu2024avatar, boer2024harnessing} to help better describes their search intents, which are overlooked by existing query expansion methods.
For example, the user query 
in Figure \ref{fig:intro}
specifies both textual requirements of the product ``highly rated'' and ``{wildlife photography}'', and relational requirement of the product ``{compatible with Nikon F-Mount lenses}''.
While existing methods may generate semantically similar expansions on this product,
they tend to make up incorrect product relations, i.e., compatibility between cameras and lenses, leading to suboptimal retrieval results.

To handle such semi-structured queries, we propose a knowledge-aware query expansion framework, augmenting LLMs with structured document relations from knowledge graphs (KG).
We first parse entities explicitly mentioned the original query with an LLM and then retrieve textual documents of these entities as well as their associated nodes and relations on KG.
While existing KG-based question answering methods \cite{yasunaga-etal-2021-qa, zhang2022greaselm, grapeqa2023} filter out irrelevant relations by scoring the node relevance based on semantic similarity between the query and entity names, e.g., ``Nikon camera'',
they overlook the rich textual details of entities queried by users, e.g., ``highly rated'' and ``wildlife''.

To address this, we leverage document texts as rich KG node representations and use document-based relation filtering to extract query-focused relations.
Then, with collected textual and relational knowledge as inputs, LLM generates query expansions that are grounded to the document corpus while preserving user-specified document relations.
The expanded queries are then utilized for the final retrieval as our \textbf{K}nowledge-\textbf{A}ware \textbf{R}etrieval (KAR).
Extensive experiments are conducted on three textual and relational semi-structured retrieval datasets in the STaRK benchmark \cite{wu2024stark} for product, 
academic paper, 
and biomedical search, respectively. 
The results show that our method outperforms state-of-the-art query expansion methods and achieves at least on par performances compared to LLM-based retrieval agent.

In summary, we make the following contributions:
i) To handle complex search queries with both textual and relational requirements, we propose a knowledge-aware query expansion framework augmenting LLMs with KG;
ii) To address the limitation of entity-based scoring, we use document texts as KG node representations and adopt document-based relation filtering for \textbf{K}nowledge-\textbf{A}ware \textbf{R}etrieval (KAR);
iii) Experiments on three semi-structured retrieval datasets show the advantages of our method and its practical applicability.

\section{Related Work}

\subsection{LLM-based Query Expansion}
Query expansion has been a widely adopted technique in information search applications \cite{azad2019query}, which expands the original query with additional contexts to match target documents.
Earlier studies use initially retrieved documents as pseudo-relevance feedback (PRF) \cite{yu2003improving, cao2008selecting, lv2010positional, li2022improving}, extracting relevant content as supplemental information.
However, the effectiveness of these methods are limited by the quality of initial retrievals.

Recently, LLM-enhanced information retrieval has been a prominent area \cite{zhu2023large}, where 
LLMs have been utilized to generate query expansions with their intrinsic knowledge.
HyDE \cite{gao-etal-2023-precise} employs an LLM to directly generate hypothetical documents that answer the query and then uses embeddings of them to retrieve similar real documents.
Query2Doc \cite{wang-etal-2023-query2doc} further improve the expansion quality by providing LLM with few-shot examples.
\citet{jagerman2023query} also explore the use of chain-of-thought as expansions.
To address the limitation that LLMs may lack domain-specific knowledge, 
\citet{shen-etal-2024-retrieval} propose retrieval-augmented retrieval (RAR) using initial retrievals as contexts for LLMs to generate query expansions.
\citet{lei-etal-2024-corpus} employ the LLM to first extract key information from initial retrievals before expanding the query.
AGR \cite{chen-etal-2024-analyze} design Analyze-Generate-Refine, a multi-step query expansion framework, to incorporate LLMs' self-refinement ability with initial retrievals as references.
Similar verification strategy is also explored in \citet{jia-etal-2024-mill}.
Despite these advances, existing methods mostly focus on textual similarities and overlook document relations.
In comparison, our knowledge-aware query expansion augments LLMs with structured document relations from KGs for handling semi-structured retrieval tasks.

\subsection{KG-Augmented LLM}
In earlier studies, language models are used to provide text embeddings to enhance graph neural networks on KG reasoning tasks \cite{feng-etal-2020-scalable, lin-etal-2021-bertgcn, yasunaga-etal-2021-qa, spillo2023combining}.
With the emergent reasoning ability of LLMs over various textual structures,
recently KG has in turn been utilized as a structured knowledge source to augment LLMs with factual or domain-specific information for more grounded reasoning and generations \cite{pan2024unifying}.
For example, Think-on-Graph \cite{sun2024thinkongraph} conducts entity and relation explorations to retrieve relevant triples for question answering.
Reason-on-Graph \cite{luo2024reasoning} retrieves reasoning paths from KGs for LLMs to conduct faithful reasoning.
HyKGE \cite{jiang2024hykge} generates hypothesis reasoning paths to be grounded on KGs for answer generation.
LPKG \cite{wang2024learning} constructs planning data from KGs for complex question answering.
There is also a recent surge in graph retrieval augmented generation \cite{peng2024graph, xu2024retrieval, he2024g, hu2024grag}, which utilizes graph data such KGs as retrieval source for more accurate and structured response generation.

Compared to these studies on LLMs with KGs, our work differs in two key aspects.
First, most prior studies focus on knowledge graph question answering, where queries are more fact-focused and answers are precisely encoded in KG. 
Our work focuses on document retrieval, where queries tend to be more descriptive and domain-specific.
Second, in prior studies KG is the sole retrieval source. 
In our work, textual documents and knowledge graph serve as a semi-structured knowledge source with information covering diverse aspects \cite{wu2024stark}. 
Such semi-structured nature in retrieval source introduce distinct challenges for effective knowledge extraction and retrieval \cite{patel2024lotus, wu2024avatar, boer2024harnessing}.

\section{Problem Definition}
\begin{figure*}[!t]
    \centering
    \includegraphics[width=0.98\linewidth]{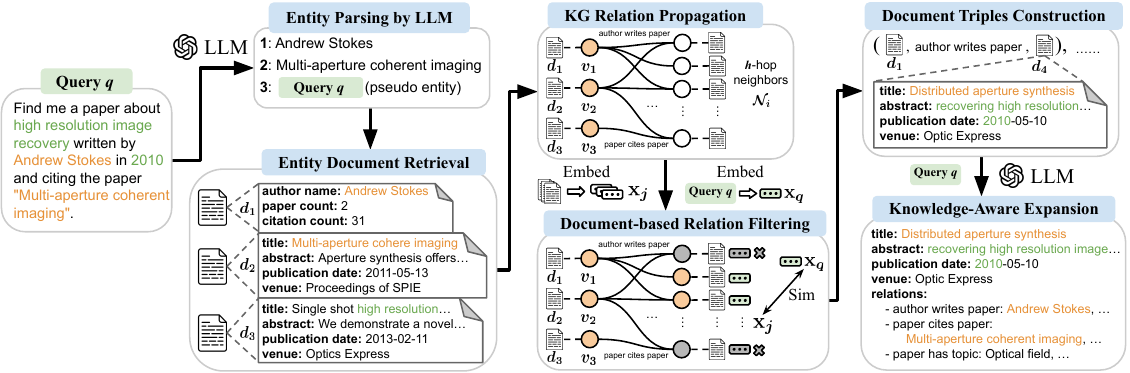}
    \caption{Overview of our knowledge-aware query expansion framework illustrated with an example academic paper search query with \textcolor[rgb]{0.39, 0.61, 0.29}{textual} and \textcolor[rgb]{0.89, 0.56, 0.22}{relational} requirements.}
    \label{fig:overview}
    \vspace{-1em}
\end{figure*}

We define our studied query expansion for textual and relational semi-structured retrieval as follows.
Follwing \citet{wu2024stark},
suppose a knowledge base contains a collection of textual documents $\mathcal{D}$ and a knowledge graph $\mathcal{G} = (\mathcal{V, R})$, where $d_i \in \mathcal{D}$ is a textual document describing an entity $i$, and $v_i \in \mathcal{V}$ is the corresponding entity node on KG, with $\mathcal{R}$ being a set of relations between different nodes.
For example, 
in a paper search scenario as Figure \ref{fig:overview}, each paper $i$ has a textual document $d_i$ that contains abstract, venue, publication date, etc., and the corresponding node $v_i$ on the knowledge graph $\mathcal{G}$ encodes its relations with other nodes such as paper citations, authorship, and field of study.

Now given a query $q$ with requirements from both unstructured texts in $\mathcal{D}$ and structured relations in $\mathcal{G}$, 
the semi-structured retrieval \cite{wu2024stark, boer2024harnessing, wu2024avatar} is to output a set of documents $\mathcal{A} \subseteq \mathcal{D}$ such that the entity described in each document satisfies both textual and relational requirements specified by query $q$.
To bridge the gap between query and documents, 
we aims to augment the original query $q$ with query expansions ${\mathcal{Q}}_e$ based on available textual and relational knowledge
as 
\begin{equation}{\mathcal{Q}}_e=f(q, \mathcal{D}, \mathcal{G}),
    \label{eq:QE}
\end{equation}
\begin{equation}
    q^{\prime} = \operatorname{Concat}(q, {\mathcal{Q}}_e),
    \label{eq:expanded_query}
\end{equation}
where $f$ represents a query expansion function and
the expansions are then appended to $q$ as the expanded query $q^\prime$ for the final document retrieval.


\section{Methodology}
In this section, we describe our knowledge-aware query expansion framework as in Figure \ref{fig:overview}.

\paragraph{Entity Parsing by LLM} As the initial step, we first utilize an LLM to extract explicitly mentioned entities from the original query $q$ given the document structures, denoted by $\mathcal{E}_q$. 
Following similar ideas in \citet{gao-etal-2023-precise}, we also consider the original query $q$ itself as a pseudo entity representing the target entity document to be retrieved,
\begin{equation}
    \mathcal{E}_q = \{q,\; \operatorname{LLM}(q)\}.
    \label{eq:entity_set}
\end{equation}
Then, entities for a paper search query may include author names, paper titles, and the query itself.

\paragraph{Entity Document Retrieval} For each mentioned entity $i \in \mathcal{E}_q$, we then use an off-the-shelf text embedding model to retrieve its associated textual document $d_i$ from $\mathcal{D}$.
As shown in Figure \ref{fig:overview}, author documents contain information of paper and citation counts while paper documents contain abstract and publication information such date and venue.

\paragraph{KG Relation Propagation}
Based on the semi-structured knowledge base, we then link each document to its corresponding entity node $v_i \in \mathcal{V}$ on KG.
For each node $v_i$, we extract based on the KG relations in $\mathcal{G}$ its $h$-hop neighbors.
We denote the set of these neighbor nodes as $\mathcal{N}_i$ and their relations to $v_i$ as the set $\mathcal{R}_i$, e.g., author writes paper, paper cites paper.
Similarly, we link each neighbor node $v_j \in \mathcal{N}_i$ to its corresponding textual document $d_j$.

\paragraph{Document-based Relation Filtering}
For a dense KG, a node might have a large amount of neighbors including nodes that are irrelevant to the query.
Existing KG-based methods \cite{yasunaga-etal-2021-qa, zhang2022greaselm, grapeqa2023} filter out irrelevant relations by scoring the relevance between nodes and queries based on entity names.
Such entity-based approach, however, overlooks the rich textual details of entities.
For example, an entity name in KG for paper search
in Figure \ref{fig:overview}
is simply a paper title, while further details of the paper, such as the abstract content ``high resolution image recovery'' and publication information ``2010'', are often not directly available in KG, despite being frequently queried by users.

To this end, we leverage the associated document texts as rich representations of KG nodes and use document-based relation filtering to get query-focused relations.
Specifically, given the set of neighbor nodes $\mathcal{N}_i$, instead of using simply the entity names, we 
embed the textual document $d_j$ for each neighbor node with a text embedding model as $\mathbf{x}_j=\operatorname{Embed}\left(d_j\right)$ and we also embed the original query $q$ using the same embedding model $\mathbf{x}_q=\operatorname{Embed}\left(q\right)$.
Then, we compute the semantic similarity of each node $v_j$ with the query $q$ and score them as
\begin{equation}
s_{j,q}=\operatorname{Sim}\left(\mathbf{x}_j, \mathbf{x}_q\right).
\end{equation}
which reflects more accurately the relevance between query and the neighbor node utilizing richer textual details besides entity name.
Based on the similarity scores, we select the top-$k$ scored nodes as query-focused neighbors
\begin{equation}
    \mathcal{N}_{i,q} = \left\{v_j \in \mathcal{N}_i \mid s_{j,q} \in \operatorname{TopK}\left(s_{q}\right)\right\},
    \label{eq:topk}
\end{equation}
and derive corresponding query-focused relations $\mathcal{R}_{i,q} \subseteq \mathcal{R}_{i}$.
Since our document-based relation filtering uses an off-the-shelf text embedding model, it does not requires any re-training as most graph neural networks do when new nodes are added to the KG, showing the scalability of our method.

\paragraph{Document Triples Construction}

With our filtered neighbors nodes and relations, instead of constructing entity-based knowledge triples like \citet{sun2024thinkongraph} and \citet{luo2024reasoning},
we further leverage the rich textual information to construct a document-based knowledge triples as in Figure \ref{fig:overview}
\begin{equation}
    \mathcal{T}_{i,q} = \left\{ (d_i, r_{i,j}, d_j) \mid v_j \in \mathcal{N}_{i,q}, r_{i,j} \in \mathcal{R}_{i,q}\right\},
\end{equation}
where $r_{i,j}$ denotes the relation on KG from node $v_i$ to node $v_j$, e.g., paper cites paper, while $d_i$ and $d_j$ are the document texts associated with $v_i$ and $v_j$ containing details of each node, e.g., paper abstract, venue, and publication date.
Such document triples not only provide rich textual details but also preserve the structured relational knowledge from KG to enhance the information accuracy. 

\paragraph{Knowledge-Aware Expansion} 
At the last step, we 
transform our document triples $\mathcal{T}_{q}$ into texts together with the original query $q$ as LLM inputs.
Leveraging its strong textual reasoning ability, we prompt the LLM to extract useful information from $\mathcal{T}_{q}$ and generate query expansions that help answer the query $q$ as
\begin{equation}
{\mathcal{Q}}_e = \operatorname{LLM}\left(q, \mathcal{T}_{q}\right),
\label{eq:generate}
\end{equation}
where we follow \citet{shen-etal-2024-retrieval} and \citet{chen-etal-2024-analyze} to sample $n$ responses through a single LLM inference and concatenate them as the final expansion appended to the original query.
The expanded query $q^{\prime}$ as defined in Equation \ref{eq:expanded_query} is then utilized for the final embedding-based document retrieval.

Throughout the query expansion, we leverage collaboratively textual documents and KG relations to achieve our \textbf{K}nowledge-\textbf{A}ware \textbf{R}etrieval (KAR).
Since our method is zero-shot, it can be applied with various off-the-shelf LLMs and text embedding models.
Besides, since our method utilizes document texts for KG node representations and thus requires no additional model finetuning, it is scalable and flexible as new documents are added to the knowledge base.

\section{Experimental Setup}

\subsection{Datasets and Metrics}
We evaluate our method on three textual and relational semi-structured retrieval datasets from the STaRK benchmark \cite{wu2024stark}:
\begin{itemize}[left=0pt]
    \item \textbf{AMAZON}: a product search dataset where textual documents for 1.0M entities are collected from Amazon reviews \cite{he2016ups} and Q\&A records \cite{mcauley2015image} and 9.4M KG relations include products viewed or purchased together, brands and colors.
    \item \textbf{MAG}: an academic paper search dataset based on obgn-papers100M \cite{hu2020open} and MAG \cite{wang2020microsoft}, where textual documents of 1.9M entities include paper title, abstract, and publication details and 39.8M KG relations include citation and authorship information.
    \item \textbf{PRIME}: a precision medicine inquiry dataset where textual documents are collected from multiple sources for about 129K entities such as disease, drug, protein and gene, and 8.1M KG relations are from PrimeKG \cite{chandak2023building}.
\end{itemize}
The statistics of datasets are shown in Table \ref{tab:stats}, where AMAZON data has richer textual information while MAG and PRIME have denser relations.
We use the official test sets of synthetic queries and leave-out sets of human-generated queries in the STaRK benchmark as well as the following evaluation metrics: \textbf{Hit@1}, \textbf{Hit@5}, Recall@20 (\textbf{R@20}), and Mean Reciprocal Rank (\textbf{MRR}).
We present further ablation results on human-generated queries to showcase the generalizability of our method for handling real-world queries.

\begin{table}[t!]
\centering
\scriptsize
\setlength{\tabcolsep}{5.5pt}
\renewcommand{\arraystretch}{1.1}
\begin{tabular}{l|rrrr}
\toprule
& \textbf{\#entities} & \textbf{\#text tokens} & \textbf{\#relations} & \textbf{avg. degree} \\
\midrule 
AMAZON & 1,035,542 & 592,067,882 & 9,443,802 & 18.2 \\
MAG & 1,872,968 & 212,602,571 & 39,802,116 &  43.5 \\
PRIME & 129,375 & 31,844,769 & 8,100,498 & 125.2 \\
\bottomrule
\end{tabular}
\renewcommand{\arraystretch}{1.0}
\vspace{-0.5em}
\caption{Statistics of textual and relational semi-structured retrieval datasets in STaRK benchmark.}
\vspace{-1em}
\label{tab:stats}
\end{table}

\begin{table*}[t!]
\centering
\fontsize{8pt}{10pt}\selectfont
\setlength{\tabcolsep}{6pt}
\renewcommand{\arraystretch}{1}
\begin{tabular}{lcccccccccccc}
\toprule
 & \multicolumn{4}{c}{\textbf{AMAZON}} & \multicolumn{4}{c}{\textbf{MAG}} & \multicolumn{4}{c}{\textbf{PRIME}} \\
\cmidrule(lr){2-5} \cmidrule(lr){6-9} \cmidrule(lr){10-13}
\textbf{Method} & \textbf{Hit@1} & \textbf{Hit@5} & \textbf{R@20} & \textbf{MRR} & \textbf{Hit@1} & \textbf{Hit@5} & \textbf{R@20} & \textbf{MRR} & \textbf{Hit@1} & \textbf{Hit@5} & \textbf{R@20} & \textbf{MRR} \\
\midrule
\multicolumn{2}{l}{\textit{Supervised Settings}} \\
DPR & 15.29 & 47.93 & 44.49 & 30.20 & 10.51 & 35.23 & 42.11 & 21.34 & 4.46 & 21.85 & 30.13 & 12.38 \\
QAGNN & 26.56 & 50.01 & 52.05 & 37.75 & 12.88 & 39.01 & 46.97 & 29.12 & 8.85 & 21.35 & 29.63 & 14.73 \\
AvaTaR & 49.87 & \textbf{69.16} & \textbf{60.57} & 58.70 & {44.36} & {59.66} & 50.63 & {51.15} & 18.44 & 36.73 & 39.31 & 26.73 \\
\midrule
\multicolumn{2}{l}{\textit{Zero-Shot Settings}}\\
Base & 39.16 & 62.73 & 53.29 & 50.35 & 29.08 & 49.61 & 48.36 & 38.62 & 12.63 & 31.49 & 36.00 & 21.41 \\
PRF & 40.07 & 60.66 & 51.24 & 49.79 & 29.04 & 47.65 & 46.69 & 37.90 & 12.46 & 28.63 & 33.04 & 20.06 \\
HyDE & 40.31 & 64.43 & 53.71 & 51.42 & 29.98 & 50.10 & 50.02 & 39.58 & 16.85 & 37.59 & 43.55 & 26.56 \\
RAR & \underline{51.52} & 66.63 & 54.63 & \underline{58.73} & 39.02 & 52.87 & 50.87 & 45.74 & 22.53 & 40.84 & 44.50 & 30.93 \\
AGR & 49.82 & 62.97 & 53.38 & 56.77 & 39.29 & 53.66 & {51.89} & 46.20 & \underline{25.85} & {44.41} & {46.63} & {35.04} \\
\textbf{KAR}$_{\mathrm{\textbf{w/o\;KG}}}$ & 43.54 & 60.29 & 51.83 & 51.80 & 31.14 & 46.75 & 46.86 & 38.88 & 18.03 & 36.27 & 42.00 & 26.84 \\
\textbf{KAR}$_{\mathrm{\textbf{w/o\;DRF}}}$ & 47.99 & 67.54 & 56.91 & 57.14 & \underline{45.44} & \underline{63.83} & \underline{58.67} & \underline{53.85} & \underline{25.85} & \underline{46.52} & \underline{48.10} & \underline{35.52} \\
\textbf{KAR} & \textbf{54.20} & \underline{68.70} & \underline{57.24} & \textbf{61.29} & \textbf{50.47} & \textbf{65.37} & \textbf{60.28} & \textbf{57.51} & \textbf{30.35} & \textbf{49.30} & \textbf{50.81} & \textbf{39.22} \\
\bottomrule
\end{tabular}
\renewcommand{\arraystretch}{1.0}
\vspace{-0.5em}
\caption{Retrieval results on test sets of synthetic search queries.}
\label{tab:retrieval_performance}
\vspace{-1.5em}
\end{table*}

\subsection{Baselines \& Variants}
We compare our KAR method with the following baselines and ablated variants in \textit{Zero-Shot} setting: 
\begin{itemize}[left=0pt]
    \item \textbf{Base}: retrieving based on the original query.
    \item \textbf{PRF}: the classic pseudo relevance feedback \cite{lv2010positional, li2022improving} approach expanding the query with top-$n$ initial retrieval results.
    \item \textbf{HyDE} \cite{gao-etal-2023-precise}: generating expansions directly with an LLM based on the original query.
    \item \textbf{RAR} \cite{shen-etal-2024-retrieval}: a retrieval-augmented retrieval approach using top-$n$ initially retrieved documents as contexts for generating query expansions with an LLM.
    \item \textbf{AGR} \cite{chen-etal-2024-analyze}: a recent method using a multi-step framework to analyze, generate, and then refine based on top-$n$ initial retrievals for expansion optimization with an LLM.
    \item \textbf{KAR}$_\mathrm{\textbf{w/o\;KG}}$: an ablated variant of our proposed KAR method without access to KG relations and thus generates expansions based solely on textual documents of retrieved entities as in Equation \ref{eq:entity_set}.
    \item \textbf{KAR}$_\mathrm{\textbf{w/o\;DRF}}$: an ablated variant of our proposed KAR method without \textbf{D}ocument-based \textbf{R}elation \textbf{F}iltering (DRF). Instead, it conducts entity-based relation filtering as in \citet{yasunaga-etal-2021-qa} and \citet{zhang2022greaselm} using entity names.
    
\end{itemize}
For more comprehensive comparisons, we report results of some \textit{Supervised} baselines from the STaRK benchmark, including \textbf{DPR} \cite{karpukhin-etal-2020-dense} as a representative dense retrieval method, \textbf{QAGNN} \cite{yasunaga-etal-2021-qa} as a representative language model embedding-augmented graph neural network method, and the state-of-the-art LLM retrieval agent \textbf{AvaTaR} \cite{wu2024avatar}.

\subsection{Implementation Details}\label{sec:imp}
For all LLM-based query expansion methods, we use Azure OpenAI API for GPT-4o (2024-02-01) as the backbone LLM in our main experiments.
We also present additional results using LLaMA-3.1-8B-Instruct \cite{dubey2024llama} as backbone LLM in Section \ref{sec:other_model}.
Following \citet{jia-etal-2024-mill} and \citet{shen-etal-2024-retrieval}, we use the dense embeddings from OpenAI \texttt{text-embedding-ada-002} model for all query expansion methods in our main experiments, employing the dot product for similarity calculation as well as document retrieval.
We also show additional results of sparse retrieval using BM25 \cite{robertson2009probabilistic} as retriever in Section \ref{sec:other_model}.
We truncate the input when its length exceeds the context window of the backbone LLM or embedding model. All experiments run on an NVIDIA A100-SXM4-80G GPU.
The prompts for all LLM-based methods are provided in Appendix \ref{app:prompt}.

\begin{table*}[!t]
\centering
\fontsize{8pt}{10pt}\selectfont
\setlength{\tabcolsep}{6pt}
\renewcommand{\arraystretch}{1}
\begin{tabular}{lcccccccccccc}
\toprule
 & \multicolumn{4}{c}{\textbf{AMAZON}} & \multicolumn{4}{c}{\textbf{MAG}} & \multicolumn{4}{c}{\textbf{PRIME}} \\
\cmidrule(lr){2-5} \cmidrule(lr){6-9} \cmidrule(lr){10-13}
\textbf{Method} & \textbf{Hit@1} & \textbf{Hit@5} & \textbf{R@20} & \textbf{MRR} & \textbf{Hit@1} & \textbf{Hit@5} & \textbf{R@20} & \textbf{MRR} & \textbf{Hit@1} & \textbf{Hit@5} & \textbf{R@20} & \textbf{MRR} \\
\midrule
Base & 39.50 & 64.20 & 35.46 & 52.65 & 28.57 & 42.86 & 36.40 & 35.95 & 22.02 & 41.28 & 43.98 & 30.63 \\
PRF & 43.21 & 64.20 & 30.19 & 53.53 & 29.76 & 41.67 & 32.91 & 35.66 & 24.77 & 36.70 & 40.65 & 30.35 \\
HyDE & 45.68 & \underline{72.84} & {39.25} & 57.56 & 29.76 & 44.05 & {37.84} & 35.51 & 24.77 & 42.20 & 47.70 & 33.65 \\
RAR & 55.56 & {71.60} & 36.15 & 62.15 & {38.10} & {45.24} & 35.19 & {42.04} & 31.19 & 43.12 & 49.01 & 37.72 \\
AGR & 55.56 & {71.60} & 37.27 & {63.54} & 33.33 & 44.05 & 37.23 & 38.95 & {32.11} & {49.54} & {49.65} & {39.27} \\
\textbf{KAR}$_{\mathrm{\textbf{w/o\;KG}}}$ & 49.38 & 67.90 & 33.94 & 57.77 & 30.95 & 40.48 & 30.95 & 35.16 & 29.36 & 47.71 & 53.52 & 37.80 \\
\textbf{KAR}$_{\mathrm{\textbf{w/o\;DRF}}}$ & \underline{56.79} & \textbf{76.54} & \underline{39.95} & \underline{65.72} & \underline{41.67} & \underline{55.95} & \underline{44.54} & \underline{48.99} & \underline{34.86} & \underline{56.88} & \underline{56.21} & \underline{44.51}   \\
\textbf{KAR} & \textbf{61.73} & \underline{72.84} & \textbf{40.62} & \textbf{66.32} & \textbf{51.20} & \textbf{58.33} & \textbf{46.60} & \textbf{54.52} & \textbf{44.95} & \textbf{60.55} & \textbf{59.90} & \textbf{51.85} \\

\bottomrule
\end{tabular}
\renewcommand{\arraystretch}{1.0}
\vspace{-0.5em}
\caption{Retrieval results on leave-out sets of human-generated search queries.}
\vspace{-0.5em}
\label{tab:retrieval_performance_human}
\end{table*}

\begin{figure*}[t!]
    \centering
    {\includegraphics[width=0.27\textwidth]{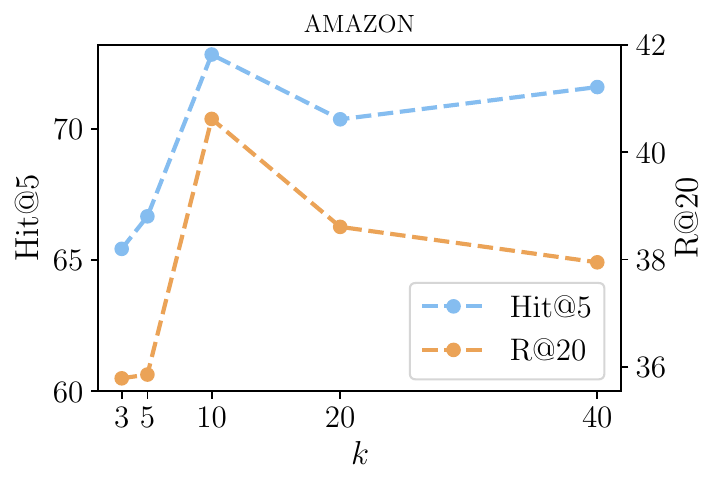}\label{fig:k_amazon}}
    {\includegraphics[width=0.27\textwidth]{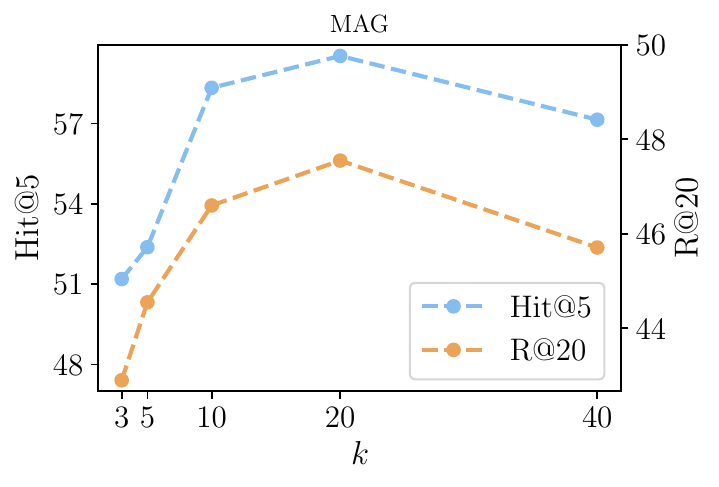}\label{fig:k_mag}}
    {\includegraphics[width=0.27\textwidth]{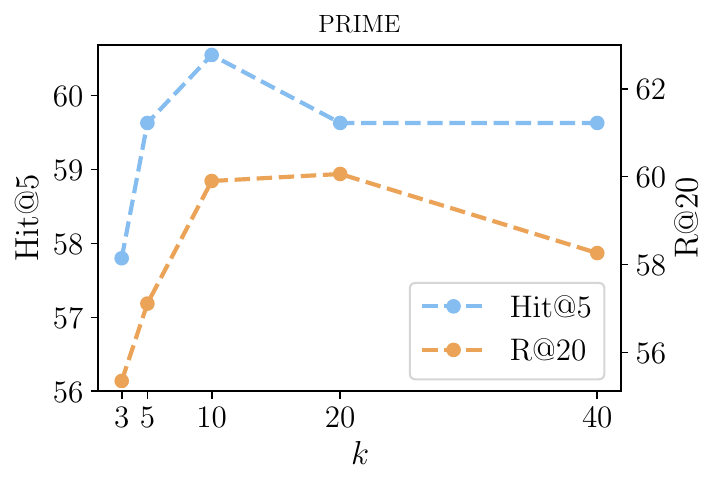}\label{fig:k_prime}}
    \vspace{-0.5em}
    \caption{Influence of different values of $k$ for filtered top-$k$ neighbors in KAR.}
    \label{fig:n_neighbor} 
    \vspace{-1em}
\end{figure*}

For hyperparameters, we set $n=3$ for PRF and all other methods utilizing the top-$n$ initial retrieval results following \citet{chen-etal-2024-analyze} and \citet{jia-etal-2024-mill}.
Existing LLM-based query expansion methods \cite{gao-etal-2023-precise, shen-etal-2024-retrieval, chen-etal-2024-analyze} usually sample multiple expansions from a single LLM inference in Equation \ref{eq:generate} to enhance the generation diversity and thus the coverage of relevant information.
Thus, we follow them and set the default number of sampled expansion generations as the same $n=3$ for a fair comparison with PRF.
The influence of different number of sampled query expansions on retrieval accuracy is further discussed in Section \ref{sec:n_generation}.
Regarding the KG parameters specifically in our KAR method, we set $h=2$ for $h$-hop neighbors following \citet{zhang2022greaselm} and \citet{grapeqa2023} to avoid exponentially increasing number of neighbor nodes farther than 2-hop.
We choose $k=10$ to select top-$k$ neighbors for query-focused relations.
The ablation results with different $k$ are presented in Section \ref{sec:n_neighbor}.
We also discuss the relative latency of compared query expansion methods in Section \ref{sec:latency}.

\section{Results}

\subsection{How does KAR perform in textual and relational semi-structured retrieval?}

We show the results on test sets of synthetic queries in Table \ref{tab:retrieval_performance} and leave-out sets of  human-generated queries in Table \ref{tab:retrieval_performance_human},
from which we observe that our KAR method achieves consistently the best or second-best performance on all metrics, validating its effectiveness for textual and relational retrieval and its generalizability to real-world scenarios.

For query expansion baselines, we find that simply using initial retrieval results as expansions, i.e., PRF, has little or even negative impact on final retrieval accuracy as low-quality initial retrievals can introduce noise for final retrievals.
Meanwhile, HyDE employs an LLM to generate query expansions directly with its intrinsic knowledge.
However, without grounded textual knowledge from the document corpus, HyDE only improves retrieval performance marginally.
For more advanced LLM-based methods, i.e., RAR and AGR, we observe that augmenting LLM with initial retrievals as contexts before expansion and utilizing its self-refinement abilities can indeed improve expansion quality and thus retrieval accuracy. 
However, the lack of relational knowledge can still lead to incorrect document relations limiting their performance on textual and relational semi-structured retrieval.

For supervised baselines, according to \citet{wu2024stark}, training challenges of encoding both textual and relational information as texts for dense retrievers and computational demands for graph neural network in QAGNN lead to significant performance gaps.
While LLM-based agent AvaTaR shows promising results after being optimized on training data, the cost and efficiency remain challenging with a high number of LLM inferences.

Moreover, compared to MAG and PRIME, we observe on AMAZON dataset higher general performance of all methods and smaller performance gaps between KAR and baselines, e.g., AvaTaR outperforms KAR on Hit@5 and R@20 metrics as shown in Table \ref{tab:retrieval_performance}.
The observation aligns well with the dataset characteristics in Table \ref{tab:stats} that AMAZON has richer textual information which can be handled better by LLMs while denser relational structures in MAG and PRIME pose challenges for LLMs to generate high quality expansions.

\begin{table*}[!t]
\centering
\fontsize{8pt}{10pt}\selectfont
\setlength{\tabcolsep}{6pt}
\renewcommand{\arraystretch}{1}
\begin{tabular}{lcccccccccccc}
\toprule
 & \multicolumn{4}{c}{\textbf{AMAZON}} & \multicolumn{4}{c}{\textbf{MAG}} & \multicolumn{4}{c}{\textbf{PRIME}} \\
\cmidrule(lr){2-5} \cmidrule(lr){6-9} \cmidrule(lr){10-13}
\textbf{Method} & \textbf{Hit@1} & \textbf{Hit@5} & \textbf{R@20} & \textbf{MRR} & \textbf{Hit@1} & \textbf{Hit@5} & \textbf{R@20} & \textbf{MRR} & \textbf{Hit@1} & \textbf{Hit@5} & \textbf{R@20} & \textbf{MRR} \\
\midrule

Base & 34.57 & 55.56 & 22.78 & 44.31 & 32.14 & 41.67 & 29.32 & 36.88 & 23.85 & 43.12 & 39.73 & 31.19 \\
PRF & 38.27 & 56.79 & 22.19 & 46.45 & 29.76 & 50.00 & 37.83 & 38.08 & 25.68 & 40.37 & 41.96 & 32.66 \\
HyDE & 36.63 & 58.44 & 23.39 & 46.29 & 32.14 & \textbf{50.40} & \textbf{39.72} & 39.84 & 28.44 & 44.65 & 44.40 & 35.79 \\
RAR & 40.74 & 59.26 & 23.39 & 49.09 & {34.23} & {50.00} & 38.41 & \underline{40.90} & 29.82 & 45.18 & \underline{45.05} & 37.02 \\
AGR & \underline{43.46} & \underline{60.25} & \textbf{24.12} & \underline{50.90} & \underline{34.76} & 48.57 & 36.90 & 40.59 & \underline{30.64} & \underline{45.50} & {44.94} & \underline{37.50} \\
\textbf{KAR} & \textbf{45.06} & \textbf{61.32} & \underline{24.10} & \textbf{52.24} & \textbf{36.51} & \underline{50.20} & \underline{38.69} & \textbf{42.41} & \textbf{32.26} & \textbf{47.25} & \textbf{46.06} & \textbf{39.14} \\
\bottomrule
\end{tabular}
\renewcommand{\arraystretch}{1.0}
\vspace{-0.5em}
\caption{Results with {BM25 as retriever} on human-generated queries.}
\vspace{-0.5em}
\label{tab:retrieval_performance_human_bm25}
\end{table*}

\begin{table*}[!t]
\centering
\fontsize{8pt}{10pt}\selectfont
\setlength{\tabcolsep}{6pt}
\renewcommand{\arraystretch}{1}
\begin{tabular}{lcccccccccccc}
\toprule
 & \multicolumn{4}{c}{\textbf{AMAZON}} & \multicolumn{4}{c}{\textbf{MAG}} & \multicolumn{4}{c}{\textbf{PRIME}} \\
\cmidrule(lr){2-5} \cmidrule(lr){6-9} \cmidrule(lr){10-13}
\textbf{Method} & \textbf{Hit@1} & \textbf{Hit@5} & \textbf{R@20} & \textbf{MRR} & \textbf{Hit@1} & \textbf{Hit@5} & \textbf{R@20} & \textbf{MRR} & \textbf{Hit@1} & \textbf{Hit@5} & \textbf{R@20} & \textbf{MRR} \\
\midrule

Base & 39.50 & 64.20 & 35.46 & 52.65 & 28.57 & \underline{42.86} & \underline{36.40} & 35.95 & 22.02 & 41.28 & 43.98 & 30.63 \\
PRF & 43.21 & 64.20 & 30.19 & 53.53 & 29.76 & 41.67 & 32.91 & 35.66 & 24.77 & 36.70 & 40.65 & 30.35 \\
HyDE & 43.21 & 65.43 & \underline{36.11} & 53.92 & 22.62 & 38.10 & 29.78 & 29.23 & 21.10 & 39.45 & 42.61 & 29.95 \\
RAR & 50.62 & 62.96 & 33.67 & 57.84 & \underline{33.33} & 41.67 & 32.01 & \underline{37.37} & \underline{28.44} & \underline{44.95} & \underline{50.12} & \underline{36.13} \\
AGR & \underline{51.85} & \underline{70.37} & 35.70 & \underline{59.91} & 30.95 & 39.29 & 32.11 & 35.20 & 23.85 & \underline{44.95} & 50.00 & 34.43 \\
\textbf{KAR} & \underline{51.85} & \textbf{71.60} & \textbf{37.78} & \textbf{60.70} & \textbf{42.86} & \textbf{54.76} & \textbf{41.36} & \textbf{47.31} & \textbf{41.28} & \textbf{55.96} & \textbf{56.45} & \textbf{48.12} \\

\bottomrule
\end{tabular}
\renewcommand{\arraystretch}{1.0}
\vspace{-0.5em}
\caption{Retrieval results with {LLaMA-3.1-8B-Instruct as backbone LLM} on human-generated queries.}
\vspace{-1em}
\label{tab:retrieval_performance_human_llama}
\end{table*}

\subsection{Are KG and document-based relation filtering (DRF) really effective?}
In both Table \ref{tab:retrieval_performance} and \ref{tab:retrieval_performance_human}, we show results of two ablated variants of our method: KAR$_{\text{w/o\;KG}}$ which has no access to relational knowledge, and KAR$_{\text{w/o\;DRF}}$ which conducts entity-based relation filtering similarly as in \citet{yasunaga-etal-2021-qa} and \citet{zhang2022greaselm}.
KAR consistently outperforms these two variants except the Hit@5 metric on AMAZON in Table \ref{tab:retrieval_performance_human}. 
The results shows that document texts and relations are both necessary and effective in enhancing the retrieval accuracy especially with denser relation structures and they contribute collaboratively to KAR.
We also find that KAR$_{\text{w/o\;DRF}}$ achieves generally better performance than KAR$_{\text{w/o\;KG}}$.
We attribute this result to the fact that LLMs' intrinsic knowledge can mitigate the textual semantic gap between queries and documents to some extent while they lack more structured relational knowledge that should be derived from the KG.

\subsection{How does the number of filtered top-$k$ neighbors affect KAR?}\label{sec:n_neighbor}

To further study the effectiveness of incorporating KG relations, we show retrieval results of KAR with varying $k \in [3, 5, 10, 20, 40]$ for filtered top-$k$ neighbors in Figure \ref{fig:n_neighbor}.
From the results, we find that initially including more query-focused neighbors based on textual documents can indeed improve retrieval accuracy as more useful document relations are covered.
However, marginal improvement diminishes as $k$ gets larger, and we observe a decrease in retrieval accuracy when increasing $k$ to 40, which suggests that irrelevant neighbors have been included, introducing noisy document relations affecting LLM's query expansion quality.
Nevertheless, our KAR method performs competitively well across different choices of $k$.

\subsection{Does the number of sampled query expansions $n$ affect retrieval accuracy?}\label{sec:n_generation}

\begin{figure}[t!]
    \centering
    {\includegraphics[width=0.145\textwidth]{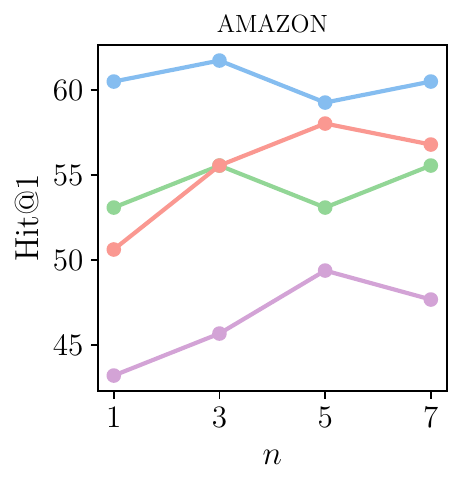}\label{fig:n_amazon}\hspace{-0.3em}}
    {\includegraphics[width=0.133\textwidth]{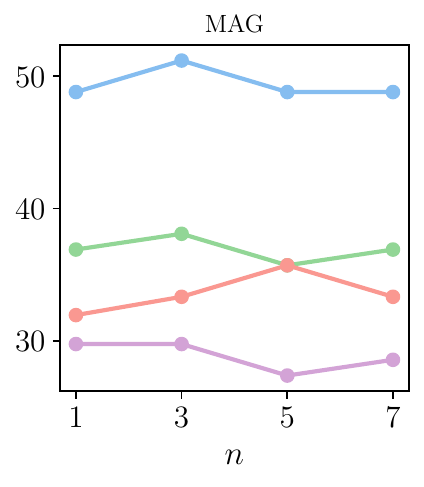}\label{fig:n_mag}\hspace{-0.3em}}
    {\includegraphics[width=0.198\textwidth]{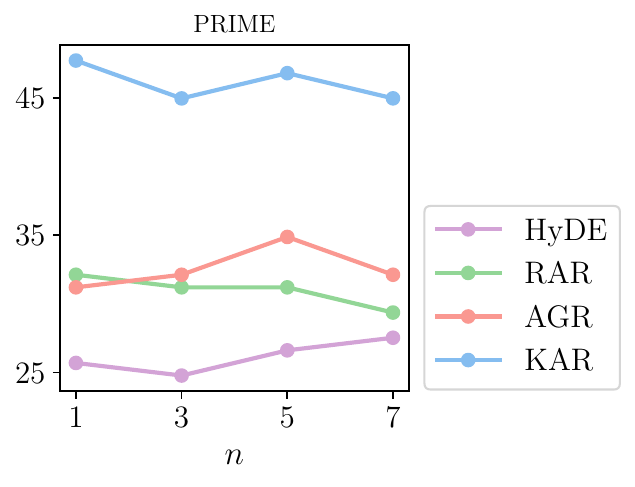}\label{fig:n_prime}}
    \vspace{-0.5em}
    \caption{Influence of sampled query expansions $n$.}
    \label{fig:n_qe} 
    \vspace{-1em}
\end{figure}

Existing LLM-based methods \cite{gao-etal-2023-precise, shen-etal-2024-retrieval, chen-etal-2024-analyze} often sample multiple expansions from a single LLM inference to enhance generation diversity and coverage. 
To study its influence on textual and relational retrieval, we show in Figure \ref{fig:n_qe} the retrieval accuracy of LLM-based methods with a varying number of sampled expansions $n \in [1, 3, 5, 7]$.
From the results, we observe that only on AMAZON do they exhibit a slightly increasing trend, while on MAG and PRIME, the number of sampled expansions does not have an obvious impact on retrieval accuracy.
We attribute this to denser relations in MAG and PRIME, where more sampled expansions from LLMs do not help in identifying structured document relations and thus cannot improve retrievals.

\subsection{{Can KAR work with other retrievers and backbone LLMs?}}\label{sec:other_model}
To further demonstrate the flexibility and scalability of KAR, we show in Table \ref{tab:retrieval_performance_human_bm25} retrieval results of all compared query expansion methods using BM25 as the sparse retriever in replacement of dense embedding-based retrieval, and in Table \ref{tab:retrieval_performance_human_llama} results with LLaMA-3.1-8B-Instruct as the backbone LLM for expansion generations.
We use the same prompts as in Appendix \ref{app:prompt} and we observe similar performance trends across different methods, with KAR consistently being the best or the second-best.
We also find that sparse retrievals based on BM25 have lower performance than dense embedding-based retrievals when augmented with query expansions, which indicates the complexity of textual and relational semi-structured retrieval.

\subsection{How does KAR compare to other methods in terms of retrieval latency?}\label{sec:latency}

\begin{figure}[!t]
    \centering
    \includegraphics[width=0.88\linewidth]{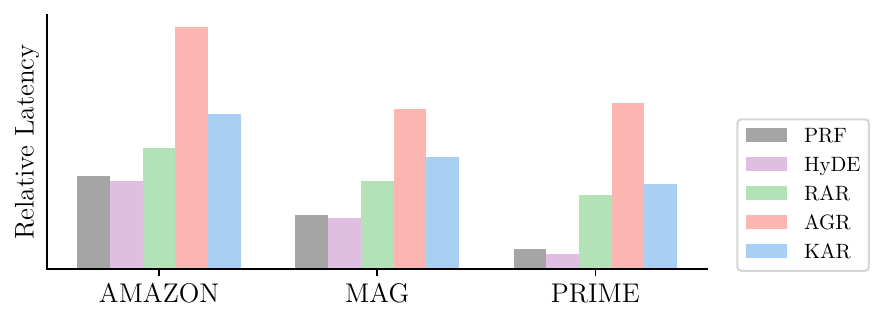}
    \vspace{-0.5em}
    \caption{Latency comparison of query expansions.
    }
    \label{fig:latency}
    \vspace{-1em}
\end{figure}

Since the latency of individual method varies based on different implementations and API versions, we keep them consistent for all methods as specified in Section \ref{sec:imp} and report results as in Figure \ref{fig:latency} for relative comparisons.
We observe that PRF and HyDE achieve the lowest latency as they only introduce one additional retrieval or one LLM inference before final retrieval.
RAR uses initial retrieval results as contexts for LLM inference, resulting in higher latency due to the additional retrieval as well as the increased textual inputs. 
AGR implements a multi-step refinement framework involving five LLM inferences, leading to about twice the latency of RAR and such observation is also suggested by \citet{chen-etal-2024-analyze}.
Instead of introducing extra LLM inferences, our KAR method employs KG to provide structured relational knowledge, allowing for fast inference, achieving considerable performance improvements while only introducing only a small amount of additional latency.

\section{Conclusion}
In this paper, we develop a knowledge-aware query expansion framework for textual and relational retrieval, utilizing KG relations between textual documents to enhance LLMs' query expansion generation.
Leveraging collaboratively textual and relational knowledge, we filter query-focused relations with document texts as rich KG node representations for our knowledge-aware retrieval.
Experiments on three semi-structured retrieval datasets of diverse domains demonstrate the advantages of our method compared against state-of-the-art query expansion methods, and showcase its applicability to handle real-world complex search queries.

\section*{Limitations}
Similar to all existing LLM-based query expansion methods, one limitation of our KAR method is the retrieval efficiency as discussed in Section \ref{sec:latency}.
While we have optimized our framework to only incorporate two LLM inferences per search query, the latency for API calls may also be influenced by varying server load.
Though we also show the effectiveness of KAR with LLaMA-3.1-8B-Instruct in Section \ref{sec:other_model} for local LLM inference, the computation constraints and deployment costs are additional important factors to be taken into consideration for practical applications.
Therefore, future works may further explore the acceleration and cost optimization of LLM inference, e.g., parallel inference, for more efficient query expansions.

\bibliography{anthology, custom}

\appendix

\newpage

\section{Prompts for LLM-based Methods}\label{app:prompt}
In this section, we provide the prompts for all LLM-based query expansion methods in our experiments.
The prompts for HyDE and RAR are presented in Table \ref{tab:prompt_1}, and Table \ref{tab:prompt_2} shows the prompts for different modules in AGR.
The prompts for our KAR method are provided in Table \ref{tab:prompt_3}.
While we generally follow the original prompts for all compared methods \cite{gao-etal-2023-precise, shen-etal-2024-retrieval, chen-etal-2024-analyze}, slight adjustments are made to adapt these methods on the evaluated textual and relational retrieval tasks.
For example, we follow \citet{wu2024avatar} to provide LLMs with the structure information of documents in each dataset in the STaRK benchmark, which helps them to generate better formatted query expansions.
The document structures of each dataset utilized in the prompts are provided in Table \ref{tab:struct}.
For more details of the datasets, please refer to \citet{wu2024stark}.

\begin{table}[h]
\centering
\small
\setlength{\tabcolsep}{4.7pt}
\begin{tabular}{lm{6cm}}
\toprule
\textbf{Method} & \textbf{Prompt} \\ \midrule
\multirow{5}{*}{HyDE} & """Given the document structures: \textcolor[rgb]{0.274, 0.671, 0.282}{\{doc\_struct\}}, write a document that answers the following user query. Return the document only without any additional text.  \\ &\\
& Query: \textcolor[rgb]{0.274, 0.671, 0.282}{\{query\}} \\& \\
& Document: """
 \\ \midrule
\multirow{12}{*}{RAR} & """Given the document structures: \textcolor[rgb]{0.274, 0.671, 0.282}{\{doc\_struct\}} and initially retrieved documents: \\& \\
& \textcolor[RGB]{250, 147, 2}{\{PRF\_doc\_1\}} \\
& \textcolor[RGB]{250, 147, 2}{\{PRF\_doc\_2\}} \\
& \textcolor[RGB]{250, 147, 2}{\{PRF\_doc\_3\}} \\& \\
& write a document that answers the following user query. Return the document only without any additional text.  \\& \\
& Query: \textcolor[rgb]{0.274, 0.671, 0.282}{\{query\}} \\& \\
& Document: """ \\ \bottomrule
\end{tabular}
\caption{Prompts for HyDE and RAR.}
\label{tab:prompt_1}
\end{table}

\begin{table}[ht]
\centering
\small
\setlength{\tabcolsep}{3.2pt}
\begin{tabular}{lm{6cm}}
\toprule
AGR & \textbf{Prompt} \\ \midrule
\multirow{5}{*}{\makecell[l]{Extract}} & """Given the following user query, write a list of keywords. Return the keywords only without any additional text.  \\& \\
& Query: \textcolor[rgb]{0.274, 0.671, 0.282}{\{query\}} \\& \\
& Keywords: """
 \\ 
\midrule
\multirow{5}{*}{\makecell[l]{Analyze}} & """Given the following user query and extracted keywords: \textcolor[RGB]{250, 147, 2}{\{extracted\_keywords\}}, do not attempt to explain or answer the question, just provide the query analysis: \\& \\
& Query: \textcolor[rgb]{0.274, 0.671, 0.282}{\{query\}} \\& \\
& Analysis: """ \\ 
\midrule
\multirow{5}{*}{\makecell[l]{Generate$^\text{1}$}} & """Given the document structures: \textcolor[rgb]{0.274, 0.671, 0.282}{\{doc\_struct\}} and the query analysis: \textcolor[RGB]{250, 147, 2}{\{query\_analysis\}}, write a document that answers the following user query. Return the document only without any additional text.  \\& \\
& Query: \textcolor[rgb]{0.274, 0.671, 0.282}{\{query\}} \\& \\
& Document: """ \\ 
\midrule
\multirow{13}{*}{\makecell[l]{Generate$^\text{2}$}} & """Given the document structures: \textcolor[rgb]{0.274, 0.671, 0.282}{\{doc\_struct\}} and initially retrieved documents: \\& \\
& \textcolor[RGB]{250, 147, 2}{\{AGR\_retrieved\_doc\_1\}} \\
& \textcolor[RGB]{250, 147, 2}{\{AGR\_retrieved\_doc\_2\}} \\
& \hspace{1.8cm}... \\
& \textcolor[RGB]{250, 147, 2}{\{AGR\_retrieved\_doc\_9\}} \\& \\
& write a document that answers the following user query. Return the document only without any additional text.  \\& \\
& Query: \textcolor[rgb]{0.274, 0.671, 0.282}{\{query\}} \\& \\
& Document: """ \\ 
\midrule
\multirow{15}{*}{\makecell[l]{Refine}} & """Given the candidate documents:\\& \\
& \textcolor[RGB]{250, 147, 2}{\{AGR\_generated\_doc\_1\}} \\
& \textcolor[RGB]{250, 147, 2}{\{AGR\_generated\_doc\_2\}} \\
& \textcolor[RGB]{250, 147, 2}{\{AGR\_generated\_doc\_3\}} \\& \\
& evaluate the accuracy and reliability of each candidate document. Identify any misinformation or incorrect facts in the answers. Then write a correct document that best answers the following user query. Return the document only without any additional text.  \\& \\
& Query: \textcolor[rgb]{0.274, 0.671, 0.282}{\{query\}} \\& \\
& Document: """ \\ 

\bottomrule
\end{tabular}
\caption{Prompts for different modules in AGR.}
\label{tab:prompt_2}
\end{table}

\begin{table}[!t]
\centering
\small
\setlength{\tabcolsep}{4pt}
\begin{tabular}{lm{6cm}}
\toprule
KAR & \textbf{Prompt} \\ \midrule
\multirow{6}{*}{Parse} & """Given the document structures: \textcolor[rgb]{0.274, 0.671, 0.282}{\{doc\_struct\}}, identify named entities in the following user query. Follow the document structures, write a document for each entity in the format: \\ & \{document type: \{document attributes\}\}.  \\ &\\
& Query: \textcolor[rgb]{0.274, 0.671, 0.282}{\{query\}} \\& \\
& Documents: """
 \\ \midrule
\multirow{12}{*}{Generate} & """Given the document structures: \textcolor[rgb]{0.274, 0.671, 0.282}{\{doc\_struct\}} and retrieved textual and relational documents: \\& \\
& \textcolor[RGB]{250, 147, 2}{\{KAR\_document\_triples\}}
 \\& \\
& extract useful information that help answer the following user query. Then, write a document that answers the following user query. Return the document only without any additional text.  \\& \\
& Query: \textcolor[rgb]{0.274, 0.671, 0.282}{\{query\}} \\& \\
& Document: """ \\ \bottomrule
\end{tabular}
\caption{Prompts for different modules in KAR.}
\label{tab:prompt_3}
\end{table}

\begin{table}[!t]
\centering
\small
\setlength{\tabcolsep}{2.3pt}
\begin{tabular}{lm{6cm}}
\toprule
\textbf{Dataset} & \textbf{Document Structures} \\ \midrule
\multirow{7}{*}{AMAZON} & \{ \\
& \hspace{0.5cm}"product": ["title", "brand", "description", \\
& \hspace{1cm}"features", "reviews", "Q\&A"], \\
& \hspace{0.5cm}"brand": ["brand\_name"], \\
& \hspace{0.5cm}"category": ["category\_name"], \\
& \hspace{0.5cm}"color": ["color\_name"] \\
& \}
 \\ \midrule
\multirow{7}{*}{MAG} & \{ \\
& \hspace{0.5cm}"paper": ["title", "abstract", "publication \\
&  \hspace{1cm} date", "venue"], \\
& \hspace{0.5cm}"author": ["name"], \\
& \hspace{0.5cm}"institution": ["name"], \\
& \hspace{0.5cm}"field\_of\_study": ["name"] \\
& \}
 \\ \midrule
\multirow{18}{*}{PRIME} & \{ \\
& \hspace{0.5cm}"disease": ["id", "type", "name", "source", \\ & \hspace{1cm}"details"], \\
& \hspace{0.5cm}"gene/protein": ["id", "type", "name", \\ & \hspace{1cm}"source", "details"], \\
& \hspace{0.5cm}"molecular\_function": ["id", "type", "name" \\ & \hspace{1cm}"source"], \\
& \hspace{0.5cm}"drug": ["id", "type", "name", "source", \\ & \hspace{1cm}"details"], \\
& \hspace{0.5cm}"pathway": ["id", "type", "name", "source", \\ & \hspace{1cm}"details"], \\
& \hspace{0.5cm}"anatomy": ["id", "type", "name", "source"], \\
& \hspace{0.5cm}"biological\_process": ["id", "type", "name", \\ & \hspace{1cm}"source"], \\
& \hspace{0.5cm}"cellular\_component": ["id", "type", "name" \\ & \hspace{1cm}"source"], \\
& \hspace{0.5cm}"exposure": ["id", "type", "name", "source"] \\
& \}
\\ \bottomrule
\end{tabular}
\caption{Document structures of the three datasets.}
\label{tab:struct}
\end{table}

\end{document}